\documentclass[review]{elsarticle}

\usepackage{lineno,hyperref}
\usepackage{amsthm}
\usepackage{amsfonts} 
\usepackage{hyperref}
\usepackage{cleveref}

\modulolinenumbers[5]

\journal{Journal of Nonlinear Analysis: Hybrid Systems}

\begin{document}

\begin{frontmatter}

\title{A General Scheme Implicit Force Control for a Flexible-Link
Manipulator
}

\author[First,Thirth]{C. Murrugarra \corref{mycorrespondingauthor} }
\ead{cmurrugarra@unbosque.edu.co}
\author[Second,Thirth]{O. De Castro}
\author[Thirth]{J.C. Grieco}
\author[Thirth]{G. Fernandez}
\address[First]{El Bosque University, Electronic Enginering Program,  Bogot\'a D.C., Colombia.}
\address[Second]{San Buenaventura University, Systems Engineering Program, Bogot\'a D.C., Colombia. }
\address[Thirth]{Simon Bolivar University, Departament of Electronics and Circuits, Caracas, Venezuela}
\cortext[mycorrespondingauthor]{Corresponding author}

\begin{abstract}
In this paper we propose an implicit force control scheme for a one-link flexible manipulator that interact with a compliant environment. The controller was based in the mathematical model of the manipulator, considering the dynamics of the beam flexible and the gravitational force. With this method, the controller parameters are obtained from the structural parameters of the beam (link) of the manipulator. This controller ensure the stability based in the \emph{Lyapunov Theory}. The controller proposed has two closed loops: the \emph{inner loop} is a tracking control with gravitational force and vibration frequencies compensation and the \emph{outer loop} is a implicit force control. To evaluate the performance of the controller, we have considered to three different manipulators (the length, the diameter were modified) and three environments with compliance modified. The results obtained from simulations verify the asymptotic tracking and regulated in position and force respectively and the vibrations suppression  of the beam in a finite time.
\end{abstract}


\end{frontmatter}

\newtheorem{thm}{Theorem}
\newtheorem{lem}[thm]{Lemma}
\newdefinition{rmk}{Remark}
\newproof{pf}{Proof}
\newproof{pot}{Proof of Theorem \ref{thm2}}

\linenumbers
\section{Introduction}
\label{sec:introduction} In the feedback control theory two types
of controllers can be identified: unconstrained and constrained
controllers. The unconstrained controller is used when the
end-effector is not in contact with the environment, for example
in robotics: feedback control for regulated and tracking control
for position and velocity the end-effector respectively. The
constrained controller is used when the end-effector is in contact
with the environment,  the force controller is classified inside
constrained controller for robotics. The applications in control
of force from manipulators have a combination of the two types of
controllers, because is necessary first to localize the
end-effector of the manipulator in the workspace in a point
desired  and then regulate to the force desired.\\ The following a
review of the state of the art in force control. The hybrid
controller proposed by Raibert \& Craig in
\cite{raibert-craig1981} and \cite{yoshikawa1986} is based on the
workspace orthogonal decomposition in two subspaces: position
control and force control. In \cite{yoshikawa} the system dynamics
was included into the position-force controller. The impedance
control by Hogan \cite{hogan} combines both, position and force
signals used in the complete \emph{manipulator-environment}
interaction. Such controllers can be used when the manipulator is
in contact with the environment and also when it's not in contact
with the environment. The explicit force control \cite{whitney}
uses a \emph{force-error}  to regulated the closed loop. The
implicit force control uses a tracking controller in stationary
state to regulate the force applied to
the environment.  \\
In this paper we propose a scheme of  Implicit Force Control for a
one-link flexible manipulator, where the end-effector interacts
with a compliant environment in the $x-y$ plane or vertical plane.
The control scheme has two closed loop controllers. The
\emph{inner loop} is a tracking controller with gravity and
vibration frequencies compensation. The \emph{outer loop} is a
implicit force controller. The scheme of  force control this based
on a dimensional finite mathematical model of the manipulator
\cite{murrugarra-grieco-TD-paper1},
\cite{murrugarra-grieco-TD-paper2}.
This paper describes: 1) The mathematical model of the manipulator
2) The  control scheme proposed  3) The stability analysis, 4) The
results and analysis obtained and 5) Conclusions.
\section{Mathematical model}
\label{sec:matemhatical_model}
The dynamics has been modelled in the joint space, where the
system is the one-link flexible manipulator with rigid rotational
joint. The gravitational force and the constrained environment are
considered in this model.
\subsection{Assumptions}
\begin{figure}[h]
\begin{center}
\includegraphics[width=9.0cm]{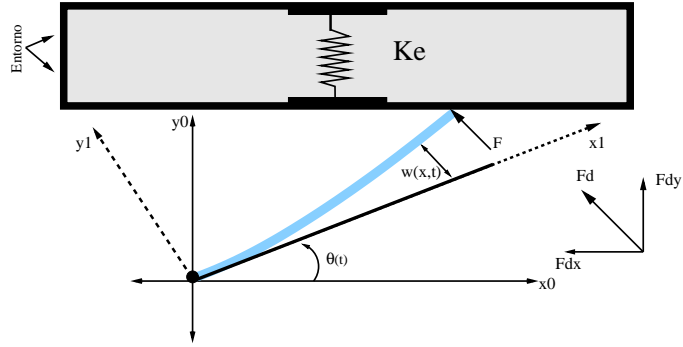} \caption{Flexible manipulator and compliant environment.}
\label{flexible_entorno}
\end{center}
\end{figure}

The development mathematical model was based in the following
assumptions: 
\begin{enumerate}
\item The dynamics of the system has been obtained from
the motion equation of \emph{Euler-Lagrange}\cite{spong2005robot}.
\item The links were
modelled as a beam \emph{Euler-Bernoulli (EB)}\citep{thomson}. 
\item The transversal
deformation was calculated in any point of the beam using the
\emph{modes-assumed} method.
\item The \emph{clamped-free} beam as
conditions of boundary of the beam.
\end{enumerate}
\begin{eqnarray}\label{eq_controlador_fuerza}
M(\wp)\ddot{\wp}+C(\wp,\dot{\wp})\dot{\wp}+g(\wp)+\eta(\wp)&=&\tau-\tau_e
\end{eqnarray}
The equation of motion of the system when the manipulator is in
contact with the environment is define by equation
(\ref{eq_controlador_fuerza}), where $M(\wp)$ the Inertia Matrix,
$C(\wp,\dot{\wp})$ the Coriolis Matrix, $g(\wp)$ the gravitational
force vector, $\eta(\wp)$ the vibrations frequency of the beam
vector, $\tau$ torque vector and $\tau_e$ $\in$ $R^{nx1}$ the
torque vector, caused by the environment as a reaction force when
the end-effector apply a force on environment. The
Fig.~\ref{flexible_entorno} show it the manipulator in contact
with environment.
\subsection{Equation}\label{ecmov_1link}
We show the  mathematical expressions of the dynamic nonlinear
model in spaces states of the manipulator. The equation
(\ref{equation-motion-eb}), represents the evolution of the system
in the time, from $L = \sum_i^n{K_i} - \sum_i^n{V_i} $, where $K_i$ is the kinetic energy for each link,
$L_i$ is the potential energy for each link, and $\wp$ is the generalized
coordinates of system corresponding to modes of vibrations of the beam. In this case $\wp=[\theta_i \quad q_{ij}]$,
where $\theta_i$ is the rotation angle in the articulation and
$q_{ij}$ is the generalized coordinate associated to temporal of
modes or vibration frequencies, $i$ is the number associated of the link, $j$ is the number of modes of
vibration flexible link, and $Q_i$ is the generalized force for each d.o.f of the system. 
\begin{eqnarray}\label{equation-motion-eb}
\frac{d}{dt}\Big{(}\frac{\partial
L}{d\dot{\wp_i}}\Big{)}-\frac{\partial L}{\partial \wp_i}= Q_i
\end{eqnarray}
We have considered the planar position of the manipulator, the
equation (\ref{position_matrix}) define the position end-effector
of the manipulator, considering the deformation of the beam in the end-effector. The kinetic energy are represent by equations (\ref{ek1}) and (\ref{ek2}), where: $A$, $\rho$, $I_b$ and $l$, are cross-sectional area, uniform mass density, inertia and length of link respectively of the beam (link).
\begin{eqnarray}\label{position_matrix}
P\left[ \begin{array}{l}
\hat{x}\\y\\\end{array} \right]^T = \left[ \begin{array}{r}
\hat{x}cos(\theta_i(t))-w_i(\hat{x},t)sin(\theta_i(t)) \\
\hat{x}sin(\theta_i(t))+w_i(\hat{x},t)cos( \theta_i(t))\\
\end{array} \right]^T
\end{eqnarray}
\begin{eqnarray}\label{ek1}
K_i=\frac{1}{2}\int_{0}^{l_{1}}\dot{P}(\hat{x})^T\dot{P}(\hat{x})dm\\
K_i=\frac{1}{2}\dot{\theta}_i^2 I_{bi} + \frac{1}{2}\rho_{i} A_{i}
\dot{\theta}_i^2 \int_{0}^{l_{i}}w_i^2(\hat{x},t) d\hat{x}+\nonumber\\
\frac{1}{2}\rho_{i} A_{i}
\int_{0}^{l_{i}}\dot{w}_i^2(\hat{x},t)d\hat{x}+
 \rho_{i} A_{i}
\dot{\theta}_i
\int_{0}^{l_{i}}\hat{x}\dot{w}_i(\hat{x},t)d\hat{x}\label{ek2}
\end{eqnarray}
The potential energy (V) has two components, the component
associated to the gravitational force (\textit{$V_g$})
and the component associated to the beam deformation
(\textit{$V_e$}). The equations (\ref{pe_1})-(\ref{pe_2})
represent the potencial energy of the system.
\begin{eqnarray}\label{pe_1}
V = V_g+V_e\\
V_g = - \int_{0}^{l_{i}}g^T \textbf{P} dm\\
V_e = \frac{1}{2}E I \int_{0}^{l_{i}}[w_i^"(\hat{x},t)]^2d\hat{x}
\end{eqnarray}
\begin{eqnarray}\label{pe_2}
V = \rho_{i} A_{i} g \frac{l_{1}^2}{2}sin (\theta_i) + \rho_{i}
A_{i} g
cos(\theta_i)\int_{0}^{l_{i}}w_1(\hat{x},t)d\hat{x}\nonumber\\+
\frac{1}{2}E I \int_{0}^{l_{i}}[w_i^"(\hat{x},t)]^2d\hat{x}
\end{eqnarray}
From (\ref{ek2}) and (\ref{pe_2}) and replace in
(\ref{equation-motion-eb}) for one link i.e. $i=1$, and using the separability principle
\cite{moorehead},  for $w_i(\hat{x},t)$, the equation of motion
might be obtained, furthermore expanded $w_i(\hat{x},t)=
\sum_{j=1}^{\nu}\phi_{ij}(\hat{x})q_{ij}(t)$ for $2$ modes
$(\nu=2)$, and representing the equation of the system by state
variables (\ref{state-var-system}), where $x$ is the state vector
and expanding for describe variables  $x=[\theta_1 \quad
\dot{\theta}_1\quad q_{11} \quad\dot{q}_{11}\quad q_{12}
\quad\dot{q}_{12}]^{T}$, and the control vector
$u=\tau-\tau_e$.
\begin{eqnarray}\label {state-var-system}
\dot{x}=f(x)+\varrho(x)u\nonumber\\
y=h(x)
\end{eqnarray}
The equation (\ref{sisecua2modas}), representing the mathematical
model in state variables, where the constants was defined by: $b_0
= a_0\rho_{1} A_{1} $, $b_1 = \rho_{1} A_{1}\frac{a^2_1}{a_0}$,
$b_2=\rho_{1} A_{1} a_1$, $b_3=\frac{a_1}{a_0}$ $b_4 =
\frac{l_{1}^2}{2}\rho_{1} A_{1}$, $b_5=a_2\rho_{1} A_{1}$ $b_6=E I
a_3$, $b_7= \frac{a_2}{a_0}$ $b_8=I_{b1}$ and  $a_0 =
\int_{0}^{l_{1}}\phi_{11}(\hat{x})^2d\hat{x}$, $a_1=
\int_{0}^{l_{1}}\phi_{11}(\hat{x})\hat{x}d\hat{x}$, $a_2 =
\int_{0}^{l_{1}}\phi_{11}(\hat{x})d\hat{x}$, $a_3 =
\int_{0}^{l_{1}}\Big[\frac{d\phi_{11}^2{\hat{x}}}{d\hat{x}^2}\Big]^2d\hat{x}$,
is important remarking that the equation (\ref{sisecua2modas}) can
be expanded for $n$ modes of vibrations, for details see
\cite{murrugarra-grieco-TD-paper1}.
{\small
\begin{eqnarray}
\dot{x}_1&=&x_2\nonumber\\
\dot{x}_2&=&
\Big{[}\tau-x_3[2b_{01}x_2x_4+b_{21}(x^2_2+b_{71}gc_1)-b_{31}b_{61}+{}\nonumber\\
&&{}b_4gc_1+b_{51}gs_1]-x_5[x_2(2b_{02}x_6+b_{22}x_2)+{}\nonumber\\
&&{}
b_{62}b_{32}+b_{52}gs_1]+b_{22}b_{72}gc_1\Big{]}\Big{[}b_{01}x^2_3+b_{11}+{}\nonumber\\
&&{}b_{02}x^2_5+b_{12
}+b_8\Big{]}^{-1}\nonumber\\
\dot{x}_3&=&x_4\nonumber\\
\dot{x}_4&=&\Big{[}x^2_2x_3-b_{31}\dot{x}_2-b_{71}gc_1-\frac{b_{61}}{b_{01}}x_3\Big{]}\nonumber\\
\dot{x}_5&=&x_6\nonumber\\
\dot{x}_6&=&\Big{[}x^2_2x_5-b_{32}\dot{x}_2-b_{72}gc_1-\frac{b_{62}}{b_{02}}x_5\Big{]}\label{sisecua2modas}
\end{eqnarray}
}
\section{Control Scheme}\label{sec:3}
\begin{figure}[h]
\centering
\includegraphics[width=12cm]{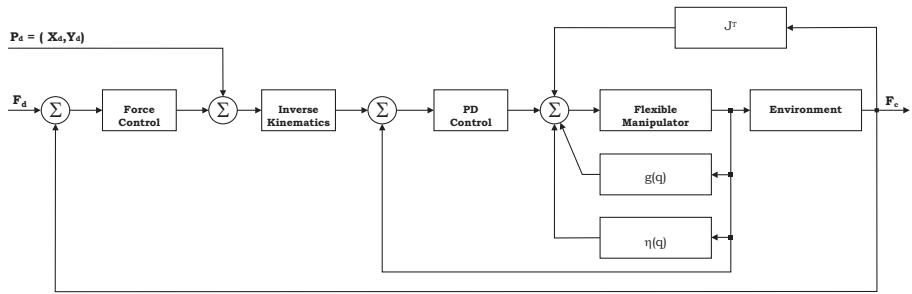}
\caption{Implicit Force Controller for the flexible-link
manipulator.} \label{esquema_ctrol_fuerza_implicito_flexible}
\end{figure}

We propose a control scheme with two closed-loop. The  Fig.
\ref{esquema_ctrol_fuerza_implicito_flexible} show this scheme.
The \emph{inner loop} is a tracking control with gravity and
vibration frequencies compensation. The \emph{outer loop} is a
force controller. The control scheme force transforms the force
error into a position difference in the components planar $x$ and
$y$. This difference is added to the reference of the tracking
controller \emph{inner loop}. The environment has been modelled as
a deformable environment or compliant surface. When the
manipulator makes contact with the environment a reaction force is
generated $Fc=(Fc_x,Fc_y)$ and this  components are feedback to
the force controller.
\section{Tracking Control Loop}\label{sec:4}
We propose a theorem that ensures the global asymptotic stability
of the PD (Proportional-Derivative) tracking controller, with both
gravity and vibration frequencies compensation on manipulator. The
theorem (\ref{teo_movimiento_pd_grav_vib}), calculate the
parameters of tuning of the tracking controller based on the
dynamics of the manipulator. This parameters only depend of the
structure of the beam and the spacial configuration of the
manipulator.
\begin{thm}\label{teo_movimiento_pd_grav_vib} Consider the
nonlinear open-loop system $\dot{x}=f(x)+\varrho(x)u$,
representing the rotation joint of the flexible-link manipulator,
with $x-y$ workspace, under gravity influence. In order to define
a closed-loop tracking controller using the control law:
\begin{eqnarray}u=K_p\tilde{\wp}+K_v\dot{\tilde{\wp}}+
M(\wp)[\ddot{\wp}_d+\Delta\dot{\tilde{\wp}}]+C(\wp,\dot{\wp})[\dot{\wp}_d+\Delta{\tilde{\wp}}]+g(\wp)+\eta(\wp)
\end{eqnarray}
assuming that $\wp_d$ and $\dot{\wp}_d$ as a set of vector
functions, $\ddot{\wp}_d$ is a  constant and the closed-loop
equation \textsf{system-controller} is non autonomous in the space
then
it can be assured that existence an unique equilibrium point,
located at the origin and with global asymptotic stability for
$K_p$ and $K_v$ $>0$, where $\tilde{\wp}$ and $\dot{\tilde{\wp}}$
are position  velocity error vectors. The $K_p$ and $K_v$, are the
proportional and derivative matrices and must be symmetric and
positive, furthermore $\Delta=K_v^{-1}K_p$ must be a nonsingular
matrix.
\end{thm}
\begin{pf} \ref{teo_movimiento_pd_grav_vib}
Let the desired reference position  $\wp_d$ for the
controller be a feasible trajectory, defined in the manipulator
workspace and the feedback control law given by the expression
(\ref{ley_controlador_pd_mov}):
\begin{eqnarray}\label{ley_controlador_pd_mov}
u=K_p\tilde{\wp}+K_v\dot{\tilde{\wp}}+
M(\wp)[\ddot{\wp}_d+\Delta\dot{\tilde{\wp}}]+C(\wp,\dot{\wp})[\dot{\wp}_d+\Delta{\tilde{\wp}}]+g(\wp)+\eta(\wp)
\end{eqnarray}
where $K_p$ and $K_v$ $\in$ $\mathbb{R}^{nxn}$ are symmetric and
positive definite matrices and $\Delta=K_v^{-1}K_p$ is a
nonsingular matrix. Rewriting the control law
(\ref{ley_controlador_pd_mov}) in functions of the
$[\tilde{\wp}^{T}\quad \dot{\tilde{\wp}}]$, obtained the equation
controller (\ref{eq_pd_mov_3})
\begin{eqnarray}\label{eq_pd_mov_3}
\frac{d}{dt}\left [\begin{array}{c}
\tilde{\wp}\\
\dot{\tilde{\wp}}\end{array}\right]= \left [\begin{array}{c}
\dot{\tilde{\wp}}\\
\frac{\Big[-K_p\tilde{\wp}-K_v\dot{\tilde{\wp}}-C(\wp,\dot{\wp})[\dot{\tilde{\wp}}+\Delta\tilde{\wp}]
\Big]}{M(\wp)}\end{array}\right]
\end{eqnarray}
where (\ref{eq_pd_mov_3}) is a non autonomous differential
equation, with an equilibrium point in the origin
$[\tilde{\wp}^{T}\quad \dot{\tilde{\wp}}]=0$ $\varepsilon$ $\mathbb{R}^{2n}$.\\
The equation (\ref{ec_pd_mov_4}) will be used as the
\emph{Lyapunov} candidate function, this function is defined from
dynamics of the manipulator.
\begin{eqnarray}\label{ec_pd_mov_4}
{\bigvee}(t,\tilde{\wp},\dot{\tilde{\wp}})=[\dot{\tilde{\wp}}+\Delta{\tilde{\wp}}]^TM(\wp)[\dot{\tilde{\wp}}+\Delta{\tilde{\wp}}]+\tilde{\wp}^T
K_p\tilde{\wp}
\end{eqnarray}
According to statement of Robotics Theory, the inertia matrix
$M(\wp)$ is symmetric and positive definite, and by definition
$K_p$ is also symmetric and positive definite
\cite{murrugarra-grieco-TD-paper3}, it can be assured that
$\bigvee$ is also globally positive definite. Replacing
$K_p=K_v\Delta$ in
$[\dot{\tilde{\wp}}+\Delta{\tilde{\wp}}]^TM(\wp)$
\cite{kelly-santibanez}, we obtain:
\begin{eqnarray}\label{ec_pd_mov_7}
\dot{\bigvee}(t,\tilde{\wp},\dot{\tilde{\wp}})=-\dot{\tilde{\wp}}^T
K_v\dot{\tilde{\wp}}-\tilde{\wp}^T \Delta^T K_v\Delta\tilde{\wp}
\end{eqnarray}
Given that by definition $K_v$ is symmetric and positive definite
and $\Delta$ is a nonsingular matrix, their product is positive
definite, proving that $\dot{\bigvee}<0$ is a globally negative
definite matrix. We can conclude that the system has global
asymptotic stability in the theorem
(\ref{teo_movimiento_pd_grav_vib}), for any symmetric positive
definite matrix $K_p$ and $K_v$.
\end{pf}
\section{Force Controller}\label{sec:5}
\subsection{Force-Torque}
\label{sec:6} In order to write (\ref{eq_controlador_fuerza}), we
can suppose that the manipulator end-effector is in contact with
environment. In other hand, applying the \emph{virtual work
principle }\cite{meirovitch}, we can consider that the forces
vector applied by the manipulator on the environment can be
associated with the Jacobian (\ref{torque_env}), obtaining a
finite dimensional model when the manipulator is it contact with
the environment.
\begin{equation}\label{torque_env}
\tau_e=J(\wp)^Tf_c
\end{equation}
where the  $J(\wp)$, is the \emph{Jacobian Matrix}, that
associates the velocity vector in the joint $\dot{\wp}$ with the
velocity in the end-effector. In other words, a transformation
from angular space to cartesian space. The Jacobian used in our
expressions have been calculated directly of the end-effector
position in cartesian space $P(x,y)$, considering the transversal
deformations of the beam  and $f_c$, as the contact force.
\begin{equation}\label{jacobiano}
J(\wp)=\frac{\partial{P(x,y)}}{\partial \wp}
\end{equation}
Since $P(x,y)$ is written  in terms of $(\theta_i , q_{ij})$, the
kinematics velocity equation for the end-effector will be:
\begin{eqnarray}\label{derivada_de_posicion}
\dot{P}(x,y)=J_{\theta_1}(q)\dot{\theta}_1+J_{q_{11}}(q)\dot{q}_{11}+\ldots+J_{q_{1\nu}}(q)\dot{q}_{1\nu}
\end{eqnarray}
The mathematical model for the flexible manipulator has been
developed  for two vibration modes $(\nu=2)$, with the resulting
Jacobian Matrix is:
\begin{equation}\label{Jacobiano_flexible}
\dot{P}(x,y)=J_{\theta_1}(q)\dot{\theta}_1+J_{q_{11}}(q)\dot{q}_{11}+J_{q_{1\nu}}(q)\dot{q}_{12}
\end{equation}
\begin{eqnarray}\label{Jacobiano_flexible2} \mathbf{J(\wp)}=
\left[\begin{array}{lll}
-l_1s1-w_1c1&-\phi_{11}(l_1)s1&-\phi_{12}(l_1)s1\\
-l_1c1-w_1s1&\phi_{11}(l_1)c1&\phi_{12}(l_1)c1\\
0&0&0\\
0&0&0\\
0&0&0\\
1&1&1\\
\end{array}\right]
\end{eqnarray}
\subsection{Environment Model} \label{sec:7}
The end-effector/contact-surface interaction is very difficult to
model. In this case the environment has been modelled as a
\emph{compliant} environment without friction.
\begin{equation}\label{fuerza_de_contacto}
f_c=K_e\Big[P(x,y)-P_0(x,y)\Big]
\end{equation}
The contact force $f_c$ have been representing as a position
difference between end-effector $P(x,y)$  and contact point
$P_0(x,y)$ more a $K_e$, that represent the environment stiffness
coefficient. Since we consider a compliant environment, can be
represent as a constant symmetric positive definite matrix.
\subsection{Control Law}\label{sec:8}
The implicit force control scheme was constructed as the
\emph{outer loop} that associates the contact force
(\ref{fuerza_de_contacto}) with a position-velocity vector so that
the difference between the desired force $(f_d)$ and the contact
force $(f_c)$ can be translated to a position and velocity
difference $\Delta P(x,y)$ and $\Delta\dot{P}(x,y)$ respectively,
adding the last one to the reference signals of the tracking
controller \emph{inner loop}. In \cite{shutter-vanbrussel} an
implicit force regulation scheme is proposed, with an outer force
control \emph{PI}, where the control law is given by
(\ref{segunda_ley_denewton4}), being the controller proportional
contribution.
\begin{eqnarray}\label{segunda_ley_denewton2}
\dot{P}_d(x,y)=-K_e^{-1}k_f\int_{0}^{t}f(\tau)d\tau\\
\dot{P}_d(x,y)=-k\Delta
f(t)=k[f_c(t)-f_d]\label{segunda_ley_denewton3}
\end{eqnarray}
Where $k=K_e^{-1}k_f$, and integrating in time the position
reference for the force control law:
\begin{equation}\label{segunda_ley_denewton4}
P_d(x,y)=-K_e^{-1}k_f\int_{0}^{t}\Delta f(\tau)d\tau
\end{equation}
\begin{figure}[h]
\begin{center}
\includegraphics[width=9cm]{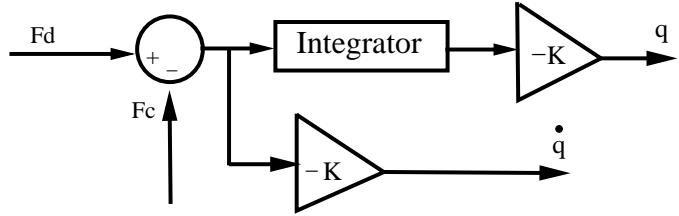}
\caption{Force/position-velocity relationship scheme control}
\label{esquema_relacion_fuerza}
\end{center}
\end{figure}
\subsection{Stability} \label{sec:9}
In order to ensure the controller stability, we have defined our
force controller based on (\ref{segunda_ley_denewton2}) and
(\ref{segunda_ley_denewton3}), defining the velocity and position
of the \emph{outer loop} . Considering the desired force $f_d$ as
a constant, the controller (\ref{segunda_ley_denewton2}) ensure an
asymptotically exact regulation, while the \emph{inner loop}
provides an asymptotically exact tracking. The inner velocity
loops with bounded errors can be seen as:
\begin{equation}
\quad \lim_{t\rightarrow\infty}sup\|\Delta
f(t)\|\leq\frac{1}{k_f}S
\end{equation}
%
If $\lim_{t\rightarrow\infty}sup\|\Delta \dot{x}(t)\|\leq S$ and
$S$ $\epsilon$ $\mathbb{R}^n$, then $0\leq S\leq\infty$ and
$k_f>0$
\begin{equation}\label{estabilidad_spr2}
\lim_{t\rightarrow\infty}sup\|\Delta \dot{x}(t)\|\leq S
\end{equation}
\section{Results and analysis}
\begin{figure}[p]
\begin{tabular}{cc}
(a)&(b)\\
\includegraphics[width=5.8cm]{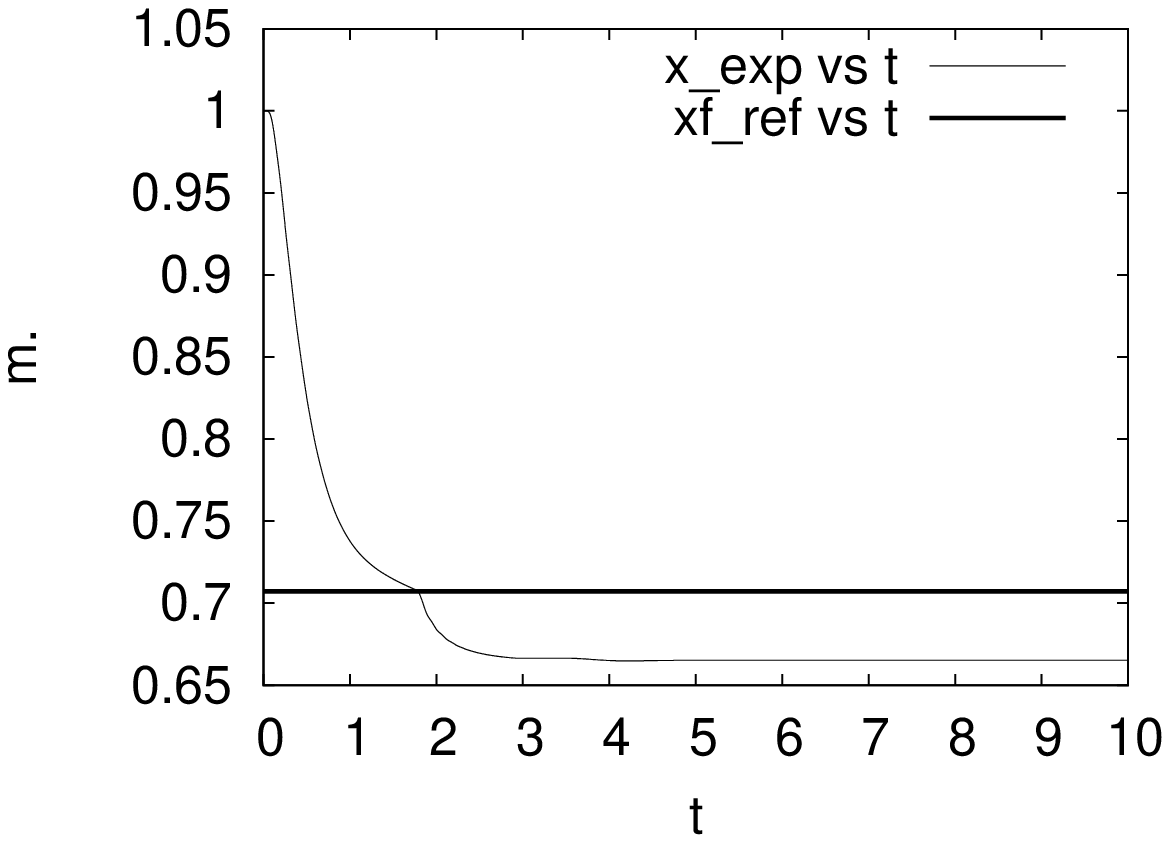}&
\includegraphics[width=5.8cm]{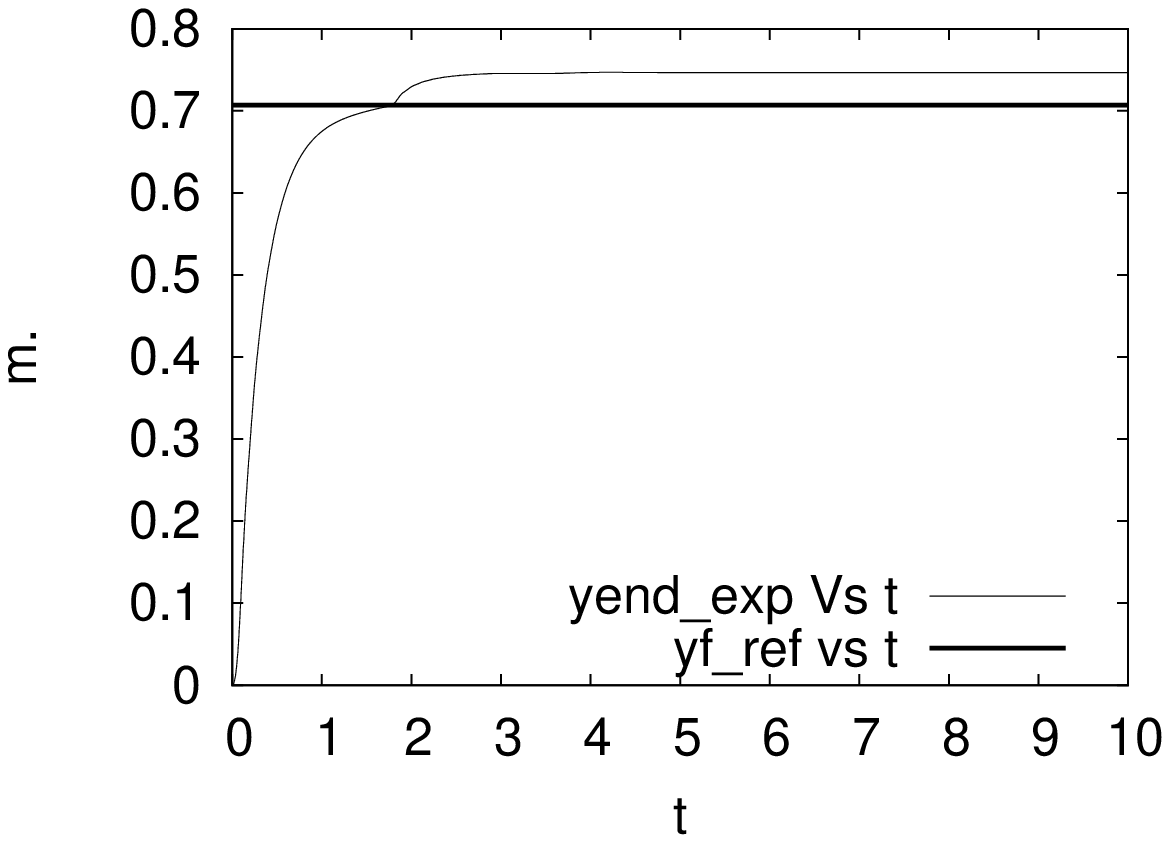}\\
(c)&(d)\\
\includegraphics[width=5.8cm]{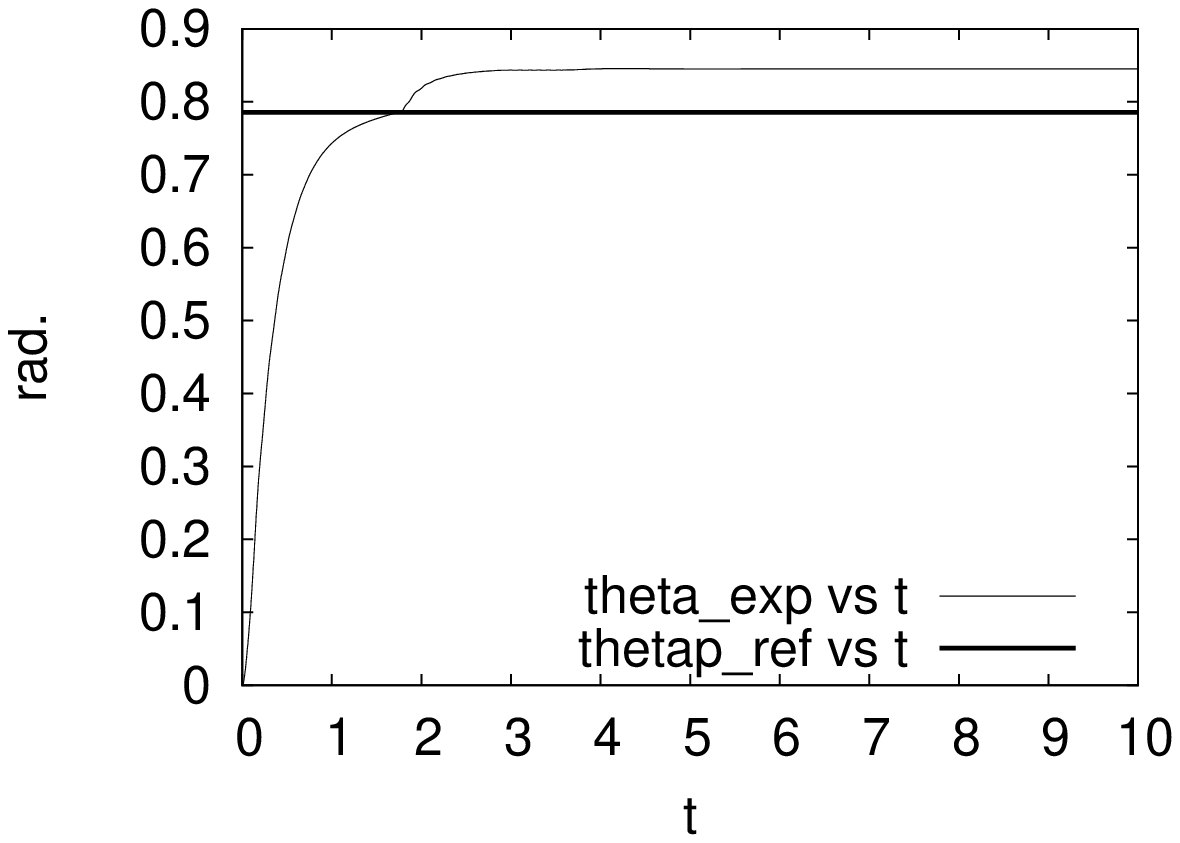}&
\includegraphics[width=5.8cm]{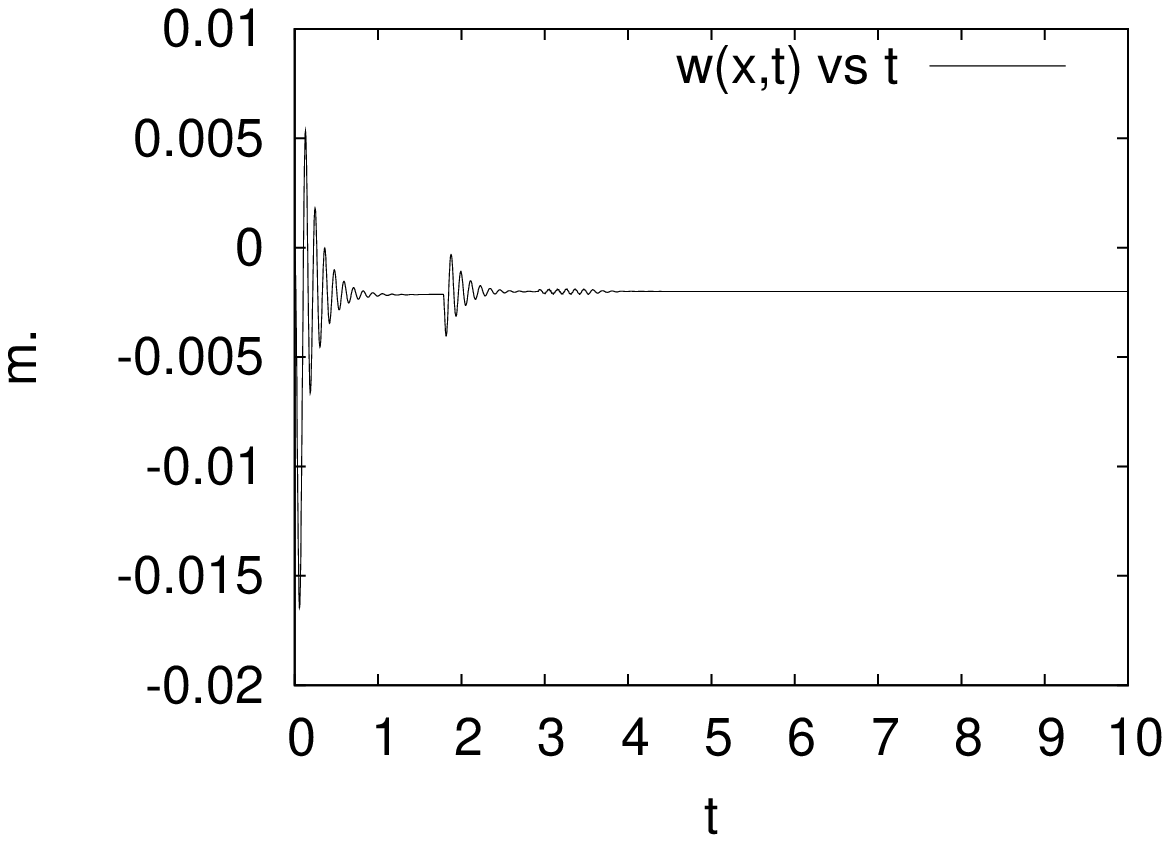}\\
(e)&(f)\\
\includegraphics[width=5.8cm]{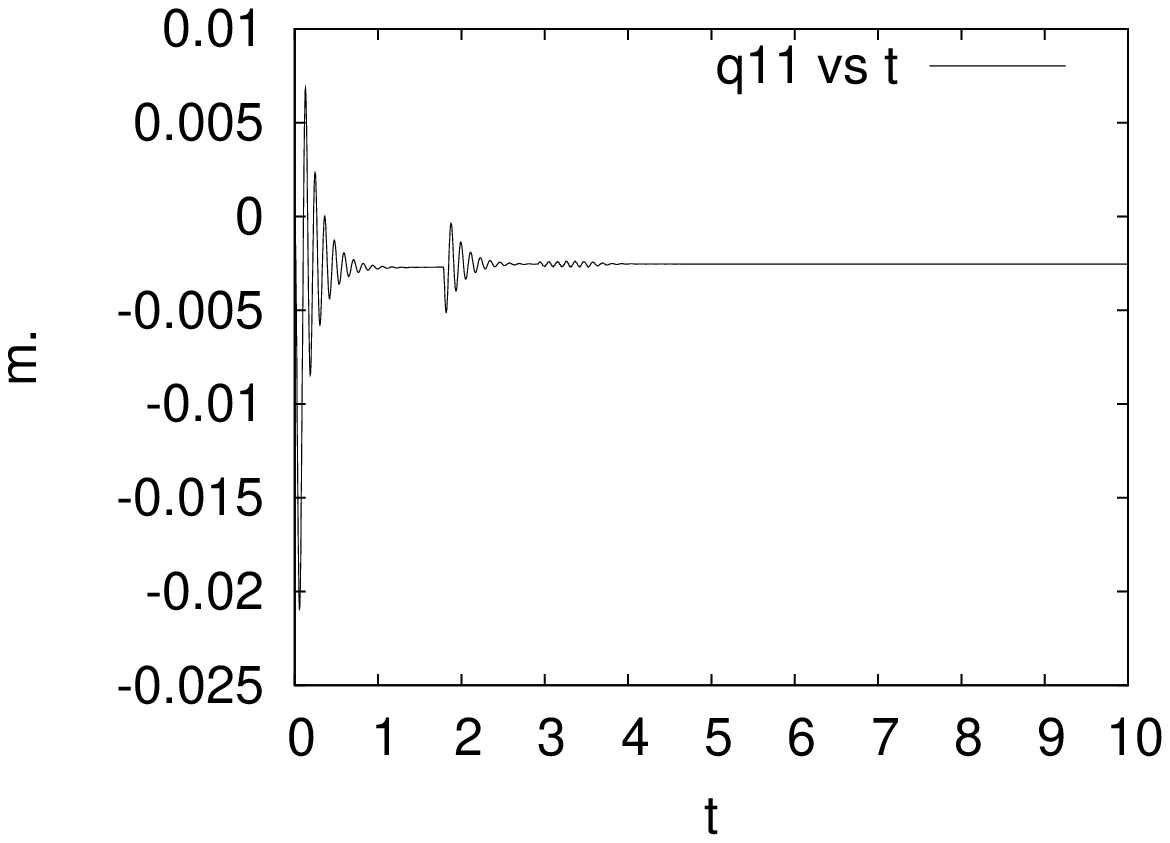}&
\includegraphics[width=5.8cm]{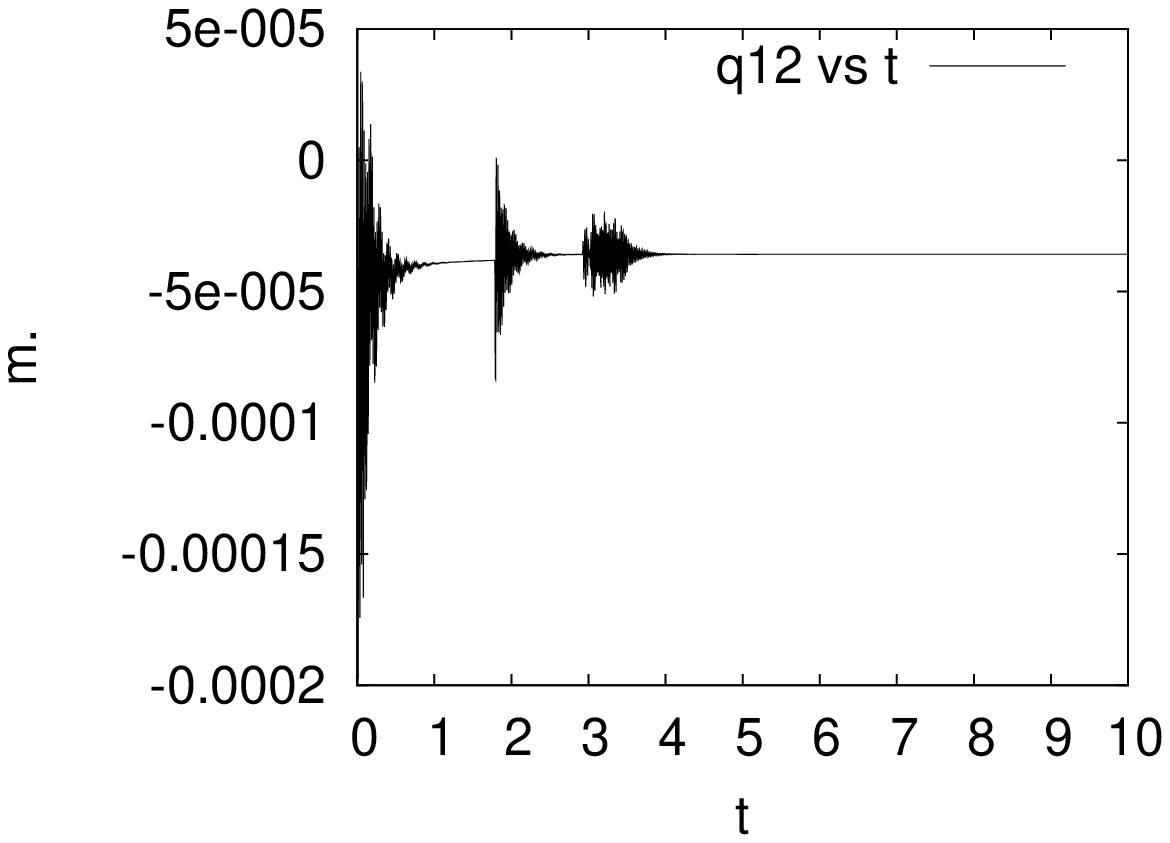}\\
\end{tabular}
\caption{Results of the Implicit Force Controller:
$Environment=86.9 N/m$, $fd=5N$.(a) Cartesian position
end-effector in the axis $x$. (b) Cartesian position end-effector
in the axis $y$  (c) Joint Position $\theta(t)$. (d) Transversal
deformation $w(\hat x, t)$. (e) First frequency of deformation E-B
beam. (f) Second frequency of deformation E-B beam.}
\label{imp_env86_9k10d01f5_1}
\end{figure}

\begin{figure}[p]
\begin{center}
\begin{tabular}{c}
\includegraphics[width=5.8cm]{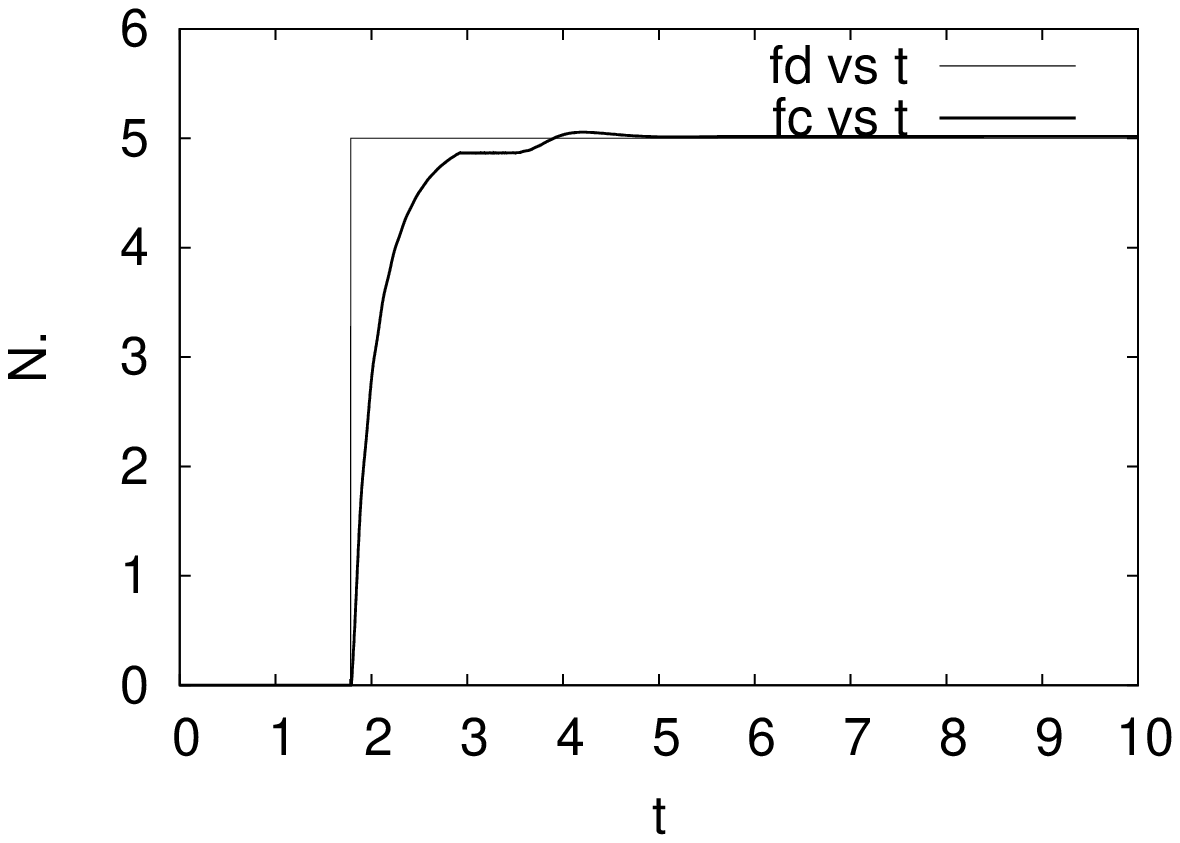}\\
(a)\\
\includegraphics[width=5.8cm]{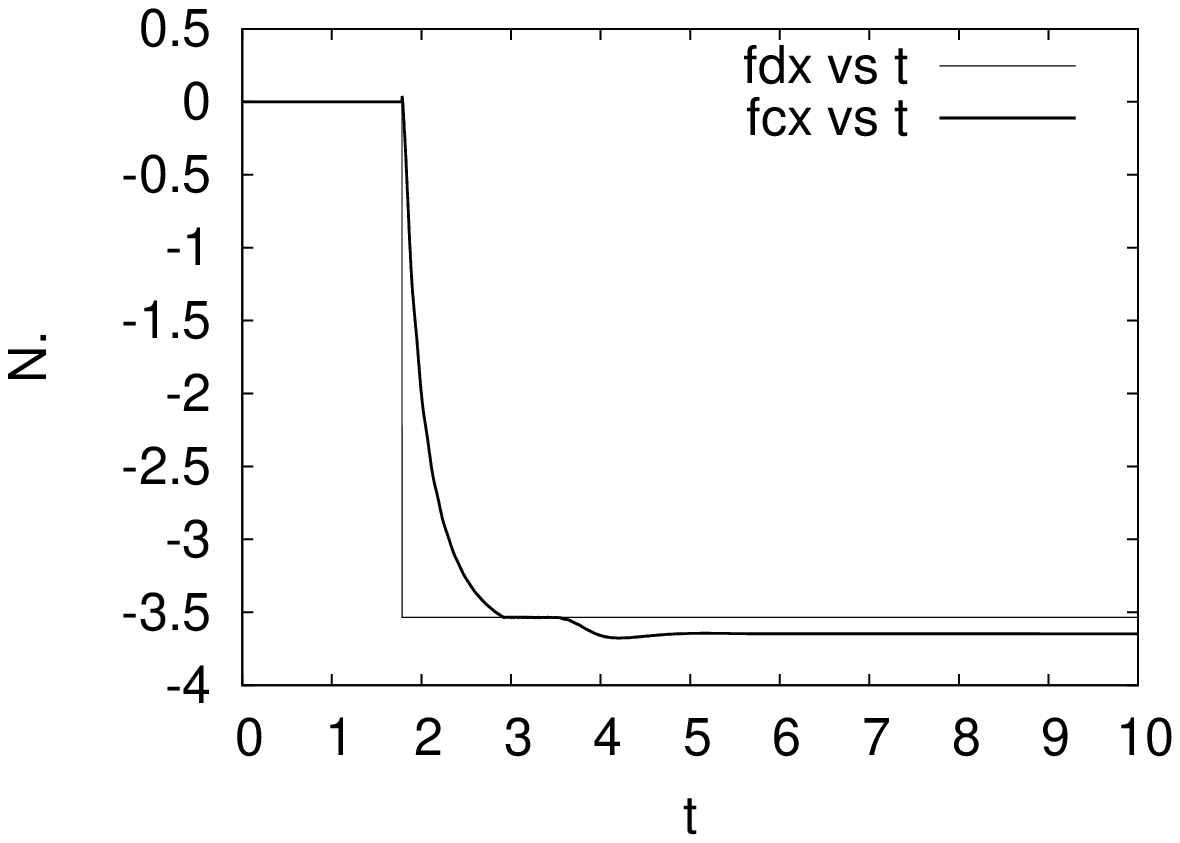}\\
(b)\\
\includegraphics[width=5.8cm]{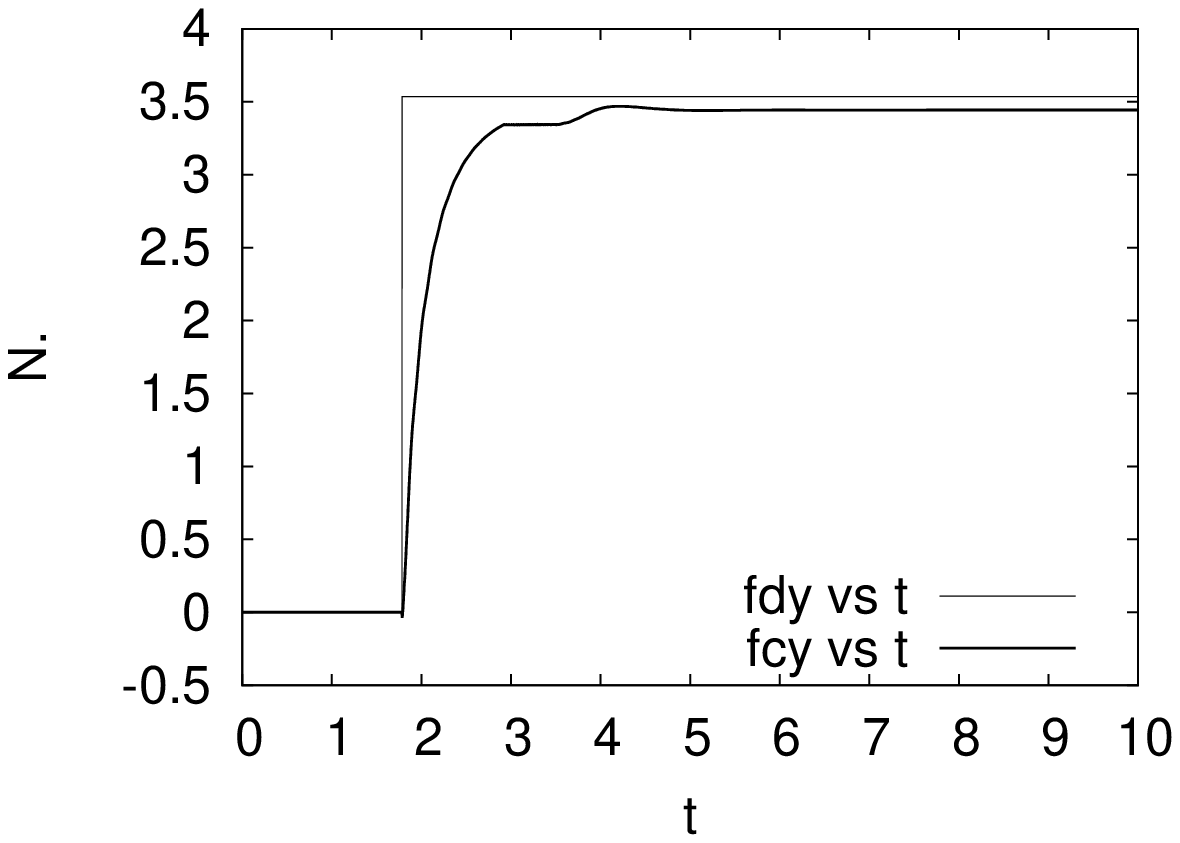}\\
(c)\\
\end{tabular}
\caption{Results of the Implicit Force Controller:
$Environment=86.9 N/m$, $fd=5N$. (a) Total Force applied to
environment. (b) Force component $F_x$ applied to the environment.
(c) Force component $F_y$ applied to the environment. }
\label{imp_env86_9k10d01f5_2}
\end{center}
\end{figure}

The proposed control scheme was tested using \emph{MatLab} and
\emph{Simulink}. The simulations have been constructed so the
manipulators end-effector was located in $P(x,y)=(0,0)m$ as the
initial position. The environment was located in
$P_e(x,y)=(0.7071,0.7071)m$, therefore the manipulator will be, at
first, under the tracking controller action, until the
end-effector makes contact with the environment. Once the contact
is made, the force controller is activated. The simulations
results are show it in the Fig. ~\ref{imp_env86_9k10d01f5_1} and
Fig. ~\ref{imp_env86_9k10d01f5_2} for the one-link flexible
manipulator. The parameters physical of the beam were: $1m.$ long
aluminum beam with cross section diameter $10^{-3}m$,
inertia $I_m=1.3254^{-06}Kgm^2$, elasticity coefficient
$EI=34.3612Nm^2$, modes-shapes $\nu=2$. The parameters of the
tracking controller are $K_p= diag[160\quad 100\quad100]$, $K_v=
diag[30\quad 1\quad 0.5]$. For the model of the  environment
$K_e=86.9N/m$ as stiffness coefficient of the environment. The
force desired is $f_d=5N$. These results show that the
end-effector reaches the final position desired applying to the
environment the desired force. We also can observe that
transversal deformations in stationary state converge to
$error=0$. 

\begin{figure}[h]
\begin{center}
\begin{tabular}{cc}
\includegraphics[width=5.7cm]{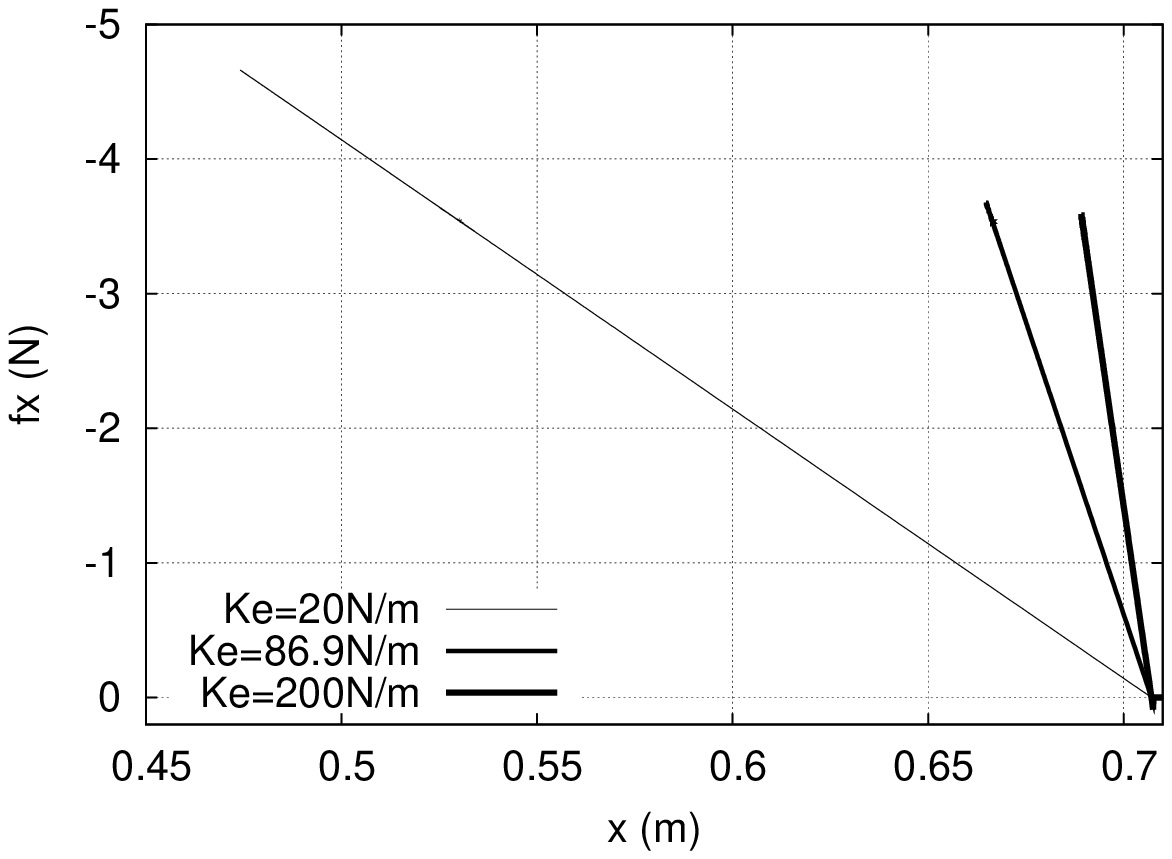}&
\includegraphics[width=5.7cm]{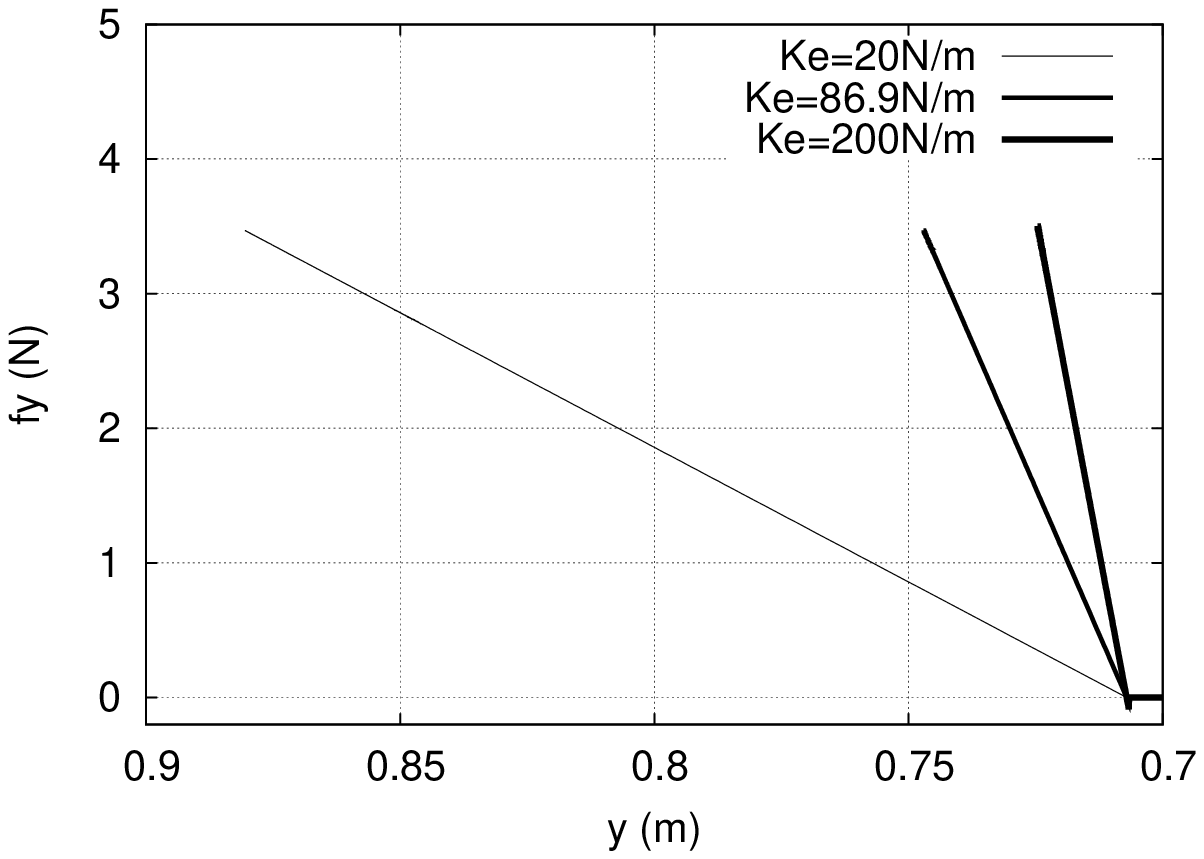}\\
\end{tabular}
\caption{Relation Force-Position for enviroments with compliance variable.} \label{relacionfuerzaposicion}
\end{center}
\end{figure}

Furthermore, in the Fig.~\ref{relacionfuerzaposicion},
we are presenting as additional results the \emph{relation
position-force}, with three environments ($K_{e1}= 20N/m$,
$K_{e2}= 86.4N/m$, $K_{e3}= 200N/m$), this results proof as the
environment is deform it, when the manipulator apply a force
(verify the environment mathematical model), visualizing this
deformation and parallel at this the convergence the both control
loops ($error _{position}$ and $error_{force}$ $\rightarrow$ 0, in
a finite time).
\section{Conclusion}
\label{sec:11} This paper proposes a general method for implement
a Implicit Force Controller based on dynamics of the manipulator.
With the control scheme proposed is possible that the manipulator
realize two works, considering indirectly the effect of the
impact. The stability analysis was based on \emph{Lyapunov
Theory}, ensure the global asymptotic stability of the control
scheme by obtaining a unique equilibrium point for controller
constants $K_p$ and $K_v$, considering the compensation  of the
gravitational force and vibrational frequencies of the beam.
Furthermore this method, can be used to prove of the resistance of
materials, because is possible know the limits of the system
(beam) associated with the vibrations amplitudes, before the deform completely when have been applied a reference force hight or when the environment is less compliant. The results were satisfactory and validate the proposed
controller.
\section*{Acknowledgments}
The authors thanks for the financial support provided by the El Bosque University, Electronic Enginering Program, San Buenaventura University, Systems Engineering Program and University Simon Bol\'{\i}var, Electronics and Circuits Department.
\bibliography{bibg}
\end{document}